\documentclass{article}

\usepackage{microtype}
\usepackage{graphicx}
\usepackage{subfigure}
\usepackage{booktabs} 

\usepackage{hyperref}

\usepackage[preprint]{icml2025}
\usepackage{amsmath}
\usepackage{amssymb}
\usepackage{mathtools}
\usepackage{amsthm}
\usepackage{algorithm}
\usepackage{algorithmic}
\usepackage{kotex}
\usepackage{graphicx}
\usepackage{subfigure}
\usepackage{booktabs} 
\usepackage{multirow}
\usepackage{amsfonts}
\usepackage[hang]{footmisc}
\usepackage{longtable}
\usepackage{pdfpages} 
\usepackage[capitalize,noabbrev]{cleveref}

\theoremstyle{plain}

\theoremstyle{definition}

\theoremstyle{remark}

\usepackage[textsize=tiny]{todonotes}

\icmltitlerunning{Adaptive Information Routing for Multimodal Time Series Forecasting}

\begin{document}

\twocolumn[
\icmltitle{Adaptive Information Routing for Multimodal Time Series Forecasting}



\icmlsetsymbol{equal}{*}

\begin{icmlauthorlist}
\icmlauthor{Jun Seo}{lair}
\icmlauthor{Hyeokjun Choe}{lair}
\icmlauthor{Seohui Bae}{lair}
\icmlauthor{Soyeon Park}{lair}
\icmlauthor{Wonbin Ahn}{lair}
\icmlauthor{Taeyoon Lim}{lair}
\icmlauthor{Junhyeok Kang}{lair}
\icmlauthor{Sangjun Han}{lair}
\icmlauthor{Jaehoon Lee}{lair}
\icmlauthor{Dongwan Kang}{lair}
\icmlauthor{Minjae Kim}{lair}
\icmlauthor{Sungdong Yoo}{lair}
\icmlauthor{Soonyoung Lee}{lair}

\end{icmlauthorlist}

\icmlaffiliation{lair}{LG AI Research, Seoul, South Korea}

\icmlcorrespondingauthor{Wonbin Ahn}{wonbin.ahn@lgresearch.ai}

\icmlkeywords{Machine Learning, ICML}

\vskip 0.3in
]



\printAffiliationsAndNotice{\icmlEqualContribution} 

\begin{abstract}
Time series forecasting is a critical task for artificial intelligence with numerous real-world applications. Traditional approaches primarily rely on historical time series data to predict the future values. However, in practical scenarios, this is often insufficient for accurate predictions due to the limited information available. To address this challenge, multimodal time series forecasting methods which incorporate additional data modalities, mainly text data, alongside time series data have been explored.
In this work, we introduce the Adaptive Information Routing (AIR) framework, a novel approach for multimodal time series forecasting. Unlike existing methods that treat text data on par with time series data as interchangeable auxiliary features for forecasting, AIR leverages text information to dynamically guide the time series model by controlling how and to what extent multivariate time series information should be combined.
We also present a text-refinement pipeline that employs a large language model to convert raw text data into a form suitable for multimodal forecasting, and we introduce a benchmark that facilitates multimodal forecasting experiments based on this pipeline.
Experiment results with the real world market data such as crude oil price and exchange rates demonstrate that AIR effectively modulates the behavior of the time series model using textual inputs, significantly enhancing forecasting accuracy in various time series forecasting tasks. 

\end{abstract}
\section{Introduction}
Time series forecasting plays a crucial role in diverse applications across finance, economics, healthcare, and industrial systems, providing essential insights for decision-making by predicting future based on historical data. Traditionally, the research community has extensively focused on unimodal forecasting methods, where the forecasting model relies solely on numerical time series data \cite{informer, autoformer, fedformer, linear, patchtst, nbeats, nhits, TSMixer, itransformer}. Although these unimodal approaches have demonstrated success in various settings, they inherently overlook the rich contextual and supplementary information that can be obtained from external modalities, limiting their capability to model complex real-world dynamics. To tackle this limitation, multimodal forecasting that integrates heterogeneous data source with time series data has emerged as a promising direction.
Inspired by the success of the Large Language Models (LLMs), recent research has begun exploring the integration of text data alongside time series data by fine-tuning LLMs as multimodal forecasting model \cite{TEMPO, Unitime, TimeLLM} or introducing the model architecture designed for multimodal forecasting \cite{TGTSF, TimeMMD}. Recent advancements in multimodal forecasting have highlighted its superior performance and robustness, demonstrating the importance of leveraging text information to improve predictive accuracy in complex scenarios.

\begin{figure}[htbp!]
    \centering
    \includegraphics[width = 0.85\columnwidth]{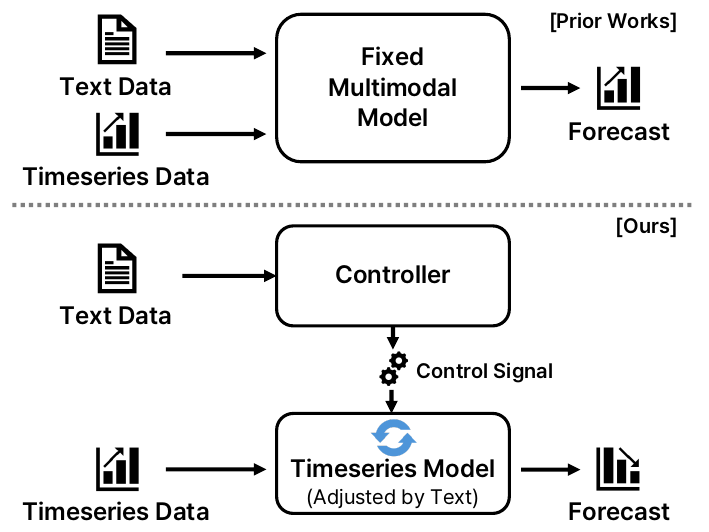}
    \caption{Comparison between the existing multimodal forecasting methods (above) and the proposed method (below)
    }
    \label{fig0}
\end{figure}

While prior works have shown remarkable achievements in multimodal time series forecasting, 
these methods typically treat text data as auxiliary inputs to the forecasting model, without adequately addressing the fundamental differences between these modalities.
Time series data typically contain local, specific, and structured information, whereas text data tend to encapsulate more global, ambiguous, and unstructured information. Ignoring these fundamental differences and treating such heterogeneous data sources equally can hinder the effective utilization of the rich information embedded in multimodal data.
To account for this fundamental distinction, we propose a novel approach that leverages text information as a controller to guide the processing of time series data within the forecasting model as shown in Figure \ref{fig0}. By doing so, the text information can influence the entire forecasting process, rather than providing information for a specific point in time and impacting only isolated moments.
Under this motivation, we introduce the Adaptive Information Routing (AIR) framework, a novel multimodal time series forecasting method that dynamically adjusts the information pathways along which time series information flows within the model based on textual cues.

Our focus is on multivariate time series forecasting, where temporal dependencies and inter-variable relationships dynamically change in response to external environmental factors.
The operation of multivariate time series forecasting model involves combining information from multiple features and multiple time points into high-level representations. The proposed AIR framework effectively controls the work of the forecasting model by controlling how and to what extent this information mixing occur reflecting the text data containing information about environmental factors. To control the information mixing, AIR decomposes the building blocks of time series forecasting models such as fully connected layers or the attention layer into two consecutive layers having latents connected to all inputs and outputs in between. By weighting the latents each of which represents a specific information pathway between inputs and outputs based on the text data, AIR realizes \textit{information routing} that controls the information mixing process of the given layer reflecting the text information. Since the information routing mechanism of AIR is applied to the layers that constitute the model rather than a specific time series forecasting model, AIR can be integrated with a various time series models to endow them with multimodal forecasting capabilities.

Our contributions are summarized as follows:
\begin{itemize}
  \item 
  We propose Adaptive Information Routing (AIR), a novel multimodal time series forecasting framework  that integrates with time series forecasting models and imparts multimodal forecasting capability by modulating the time series forecasting model based on the text information.
  \item We introduce a pipeline that refines textual inputs for multimodal time series forecasting using large language models, and employ this pipeline to build benchmarks for evaluating multimodal forecasting models.
  \item Through extensive experiment, we show that the proposed AIR framework successfully improves the performance of unimodal forecasting model by leveraging text information.
\end{itemize}


\section{Related Works}
\textbf{Time series forecasting} Time series forecasting has been an important task with various real-world applications such as resource management, financial investment, and business decision making. Traditional method of time series forecasting is to utilize statistical methods such as ARIMA \cite{arima1, arima2, arima3}. Recently, deep learning based forecasting models are showing remarkable performances in time series forecasting using various architectures such as transformers \cite{autoformer, crossformer, patchtst, itransformer, timexer} or multi layer perceptions (MLPs) \cite{nbeats, TSMixer}.

While time series forecasting models have advanced in various aspects such as performance, efficiency, and interpretability, there are inherent limitations in forecasting based solely on time series data. To make accurate forecasts of the future, it is necessary to integrate diverse sources of information, most of which are provided in forms other than time series. To overcome the limitations of time series forecasting, multimodal time series forecasting that leverage information from heterogenous data sources is necessary. Among the various types of multimodal time series forecasting, methods integrating time series and textual data have been actively investigated, driven by the recent advancements in large language models.

\textbf{Multimodal time series forecasting}
Leveraging large language models (LLMs) for time series forecasting has gained traction due to their powerful representation and sequential modeling capabilities. 
UniTime \cite{Unitime} proposes a unified multimodal forecasting framework by jointly fine-tuning an LLM with numerical data and aligned textual descriptions, significantly enhancing forecast accuracy and generalizability across diverse domains. 
TEMPO \cite{TEMPO} demonstrates that incorporating embeddings of context-rich text data alongside time series embeddings can enhance forecasting performance in the financial domain. Time-LLM\cite{TimeLLM} introduces a patch reprogramming technique that aligns time series patch embeddings to word embeddings of LLM by attention mechanism. TimeLLM also utilizes the dataset description as an auxiliary contextual information to assist time series forecasting. However, these methods utilize text data in the form of token embeddings to employ large language models as forecasting models, which results in fragmented textual information and imposes constraints on the amount of usable text data.

Complementing LLM-based methods, several recent models explicitly designed for multimodal forecasting utilize tailored multimodal architectures without relying on pretrained LLM backbones. 
TGTSF \cite{TGTSF}  utilizes textual information in the form of embedding, and integrates time series information with textual information from dynamic news and channel description by cross attention mechanism. Time-MMD \cite{TimeMMD} proposes a new comprehensive multimodal dataset that pairs diverse time series with corresponding textual data collected from report data and web search results across multiple domains. They also introduce a simple multimodal integration framework that converts the existing unimodal forecasting model into the multimodal model by independently modeling time series data and text data and combining them, and show that the proposed framework improves forecasting performance for various unimodal time series forecasting backbones. 

Despite their remarkable success, existing multimodal time series forecasting methods still have a limitation — they fail to account for the fundamental differences between time series and text data. Since text data contains relatively global information compared to time series data, text information can exert a broad and complex influence on the dynamics of time series. Therefore, multimodal forecasting models that simply use text data as an additional input face limitations in fully leveraging the information contained in text data. To tackle this limitation, we propose a novel multimodal time series forecasting framework that utilizes the text data as a controller for the time series model. 
\begin{figure}[htbp!]
    \centering
    \includegraphics[width = \columnwidth]{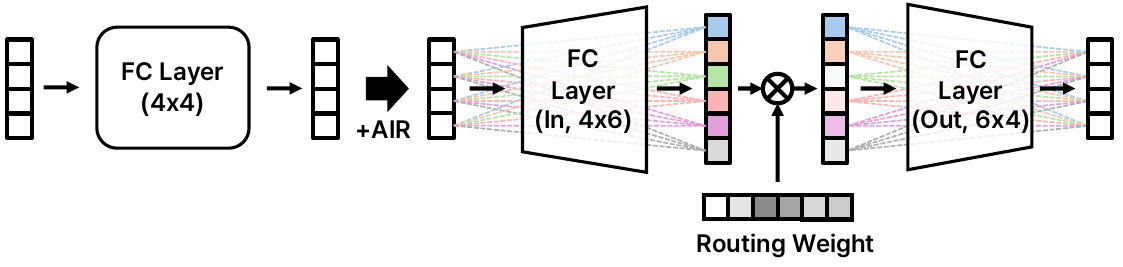}
    \caption{Adaptive Information Routing for FC layer. 
    }
    \label{fig1}
\end{figure}

\section{Method}

In this section, we present our Adaptive Information Routing (AIR) framework, designed to modulate the behavior of multivariate time series forecasting models by adjusting the information pathways within the model based on textual information.

\subsection{Adaptive Information Routing}
To illustrate our approach, we begin with the fully connected (FC) layer, a basic component of neural networks. FC layer connects each input node to each output node, with the connection strengths represented by the weights. To enable control of the operation of FC layer, we decompose the FC layer into two distinct FC layers with an intermediate latent representation. 
Figure \ref{fig1} shows how FC layer is decomposed into two distinct FC layers. Since each latent node is connected to all input and output nodes with distinct weights, each node in the latent represents a specific connection between input and output nodes. 
Therefore, we can adjust information pathways between input and output in FC layer by decomposing FC layer into two FC layers and weighting the latent nodes between them as shown in Figure \ref{fig1}. 
AIR framework leverages this decomposition structure and generates the weights applied to the latent nodes from the text embeddings. 
By adjusting the connections between input and output nodes adaptively to the text information, AIR guides the behavior of time series model based on the text information. 

\begin{figure}[h]
    \centering
    \includegraphics[width = 1.05\columnwidth]{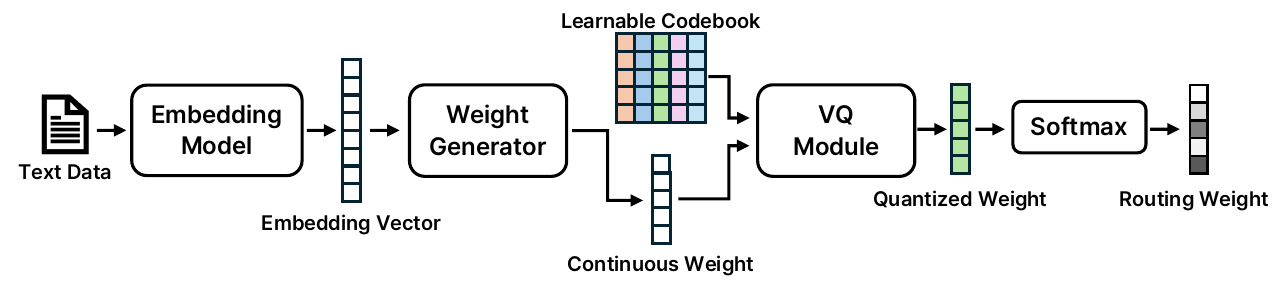}
    \caption{Architecture of information routing module. 
    }
    \label{fig2}
\end{figure}

Figure \ref{fig2} illustrates the architecture of the information routing module within the AIR framework. This module comprises three key components: the embedding model, the MLP weight generator, and vector quantization (VQ) module.
To utilize the the text information in a standardized format, we embed the text data into a fixed-size vector using a pretrained text embedding model. As our text-embedding backbone, we adopt the LGAI-Embedding-preview \cite{lgaiembedding} model, which demonstrates strong performance on the MTEB benchmark \cite{MTEB}. Based on the text embedding, the weight generator produces the weights for the latent nodes. We implement the weight generator using a simple Multi-Layer Perceptron (MLP) architecture. To stabilize the generated weights and impose a more structured organization on the latent space, we introduce a vector quantization mechanism with a learnable codebook. Instead of directly using the continuous MLP outputs, the VQ module maps each weight to its nearest codebook entry. The discrete bottleneck provided by VQ reduces noise sensitivity and organizes the routing weight space into meaningful clusters that correspond to recurring environments or market regimes. Finally, we apply a softmax activation function to the generated weights to facilitate a weighted combination across multiple paths.

This methodology can be easily extended to other architectures that learn input–output relationships through fixed weights. For example, in a convolutional neural network, a convolution layer can be decomposed into two sequential layers: an input convolution layer whose output channels correspond to the latent dimension while preserving the original input channels and kernel size, and an output convolution layer whose output channels match the original output channels but whose input channels are set to the latent dimension with a kernel size of 1. Based on this decomposition structure, the information rouging mechanism that weights the latent nodes between the two layers can be seamlessly applied to the convolutional neural network.

\begin{figure*}[h]
    \centering
    \includegraphics[width = 1.4\columnwidth]{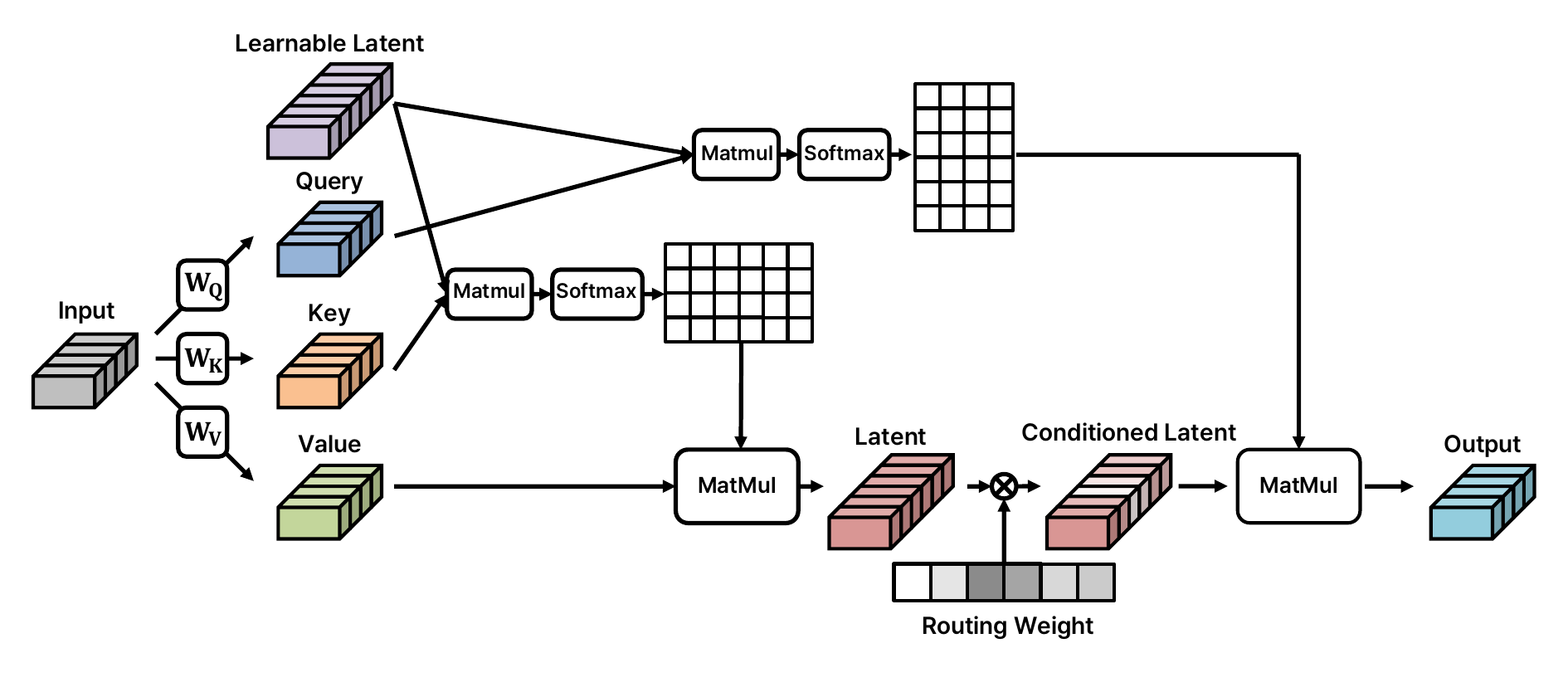}
    \caption{Adaptive Information Routing framework for attention layer. 
    }
    \label{fig3}
\end{figure*}

\subsection{Adaptive Information Routing for Attention Layer}
Recently, attention mechanism is widely used for many applications including time series forecasting.  
However, attention layers differ fundamentally from traditional architectures represented by fully connected (FC) layers.
Whereas traditional layers learn the relationships between input and output via learnable fixed weights, 
attention layers first construct an attention map and associated values, subsequently aggregating these values according to the generated attention map. 
In other words, whereas in other architectures the learned weights themselves determine which input information to combine and to what extent in order to generate the output features, in the attention layer, this role is fulfilled by the attention map generated during the forward propagation process.
Therefore, to apply the concept of information routing to the attention layer, we need to devise a method specifically tailored for the attention mechanism.

In Figure \ref{fig3},we demonstrate the integration of an adaptive information routing (AIR) framework specifically designed for attention layer. 
To implement information routing within the attention layer, adjustments on information pathways must be made directly within the attention map rather than within associated FC layers. 
To obtain latent representations each of which corresponds to a specific information pathway, we need to split the attention map into two separate attention maps: one that connects all input nodes (corresponding to each key node) to the latent representations, and another that connects the latent representations to all output nodes (corresponding to each query node). 
To facilitate this decomposition, we introduce learnable latents that function as queries in one attention map and as keys in another.
Using the learnable latents, we compute two separate attention maps: the first based on the similarity between the keys and the latents, and the second based on the similarity between the queries and the latents. The two attention maps are applied sequentially to aggregate the values, enabling us to obtain the decomposition structure with latents that are connected to all input and output nodes. The AIR framework exploits this decomposition structure and generates the routing weight from text data and applies the weight to these latents as done in FC layer, enabling the adaptive control of attention layer based on the text information. 

\section{Event-Based Text Data Refinement}
\label{section:data_pipeline}
\begin{figure*}[h]
    \centering
    \includegraphics[width = 1.8\columnwidth]{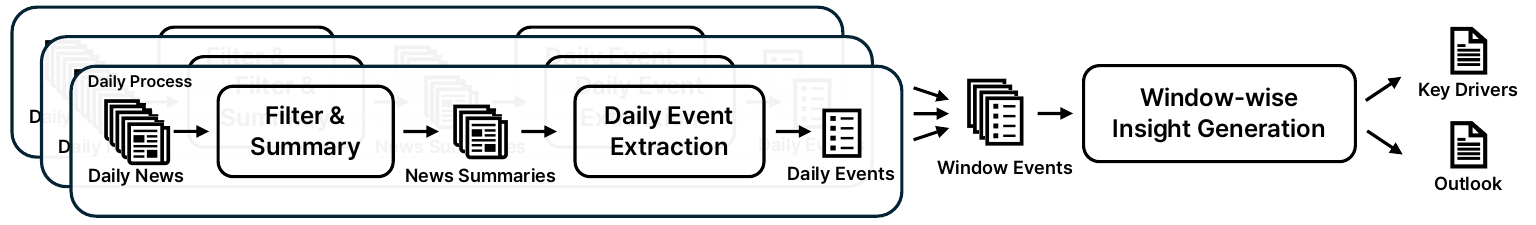}
    \caption{Event-Based Text Data Refinement Pipeline 
    }
    \label{fig_ebr}
\end{figure*}
Text data contains valuable diverse real-world signals that are not representable in time-series form but are highly useful for forecasting. Among various textual sources, news articles provide timely and continuously updated information, but they are often unrefined, noisy, and highly dispersed. Therefore, to effectively incorporate news data for forecasting, it is essential to organize, summarize, and refine the scattered information into a more coherent and informative form.
In this section, we introduce a data-refinement pipeline that employs large language models to process daily news data containing sparse and scattered information and to distill it into compact summary text that is informative for forecasting. 
To obtain organizes information from news data, our data-refinement pipeline identifies influential events across the news stream and subsequently extracts forecast-relevant information grounded in these events.

In Figure \ref{fig_ebr}, we demonstrates our Event-Based Text Data Refinement pipeline. In the first stage, we use an LLM to generate a concise, paragraph-level summary of each news article. At the same time, the LLM evaluates the article’s relevance to the forecasting domain—such as exchange rates or oil prices—and filters out unrelated content. In the second stage, we identify the top $N$ events that have the greatest impact on the forecasting domain, based on the summaries of news published on a given date. Each extracted event consists of an event name, a concise summary of the event, and a rationale explaining why the event influences the target domain. 
In the third stage, we aggregate events over multiple days, and generate two forms of textual signals based on their content: a key driver that analyzes the changes observed in each forecasting target and explains their underlying causes, and an outlook that predicts future movements in the target. Similar to time-series forecasting models that use fixed-length historical windows as input, we also group events within a temporal window to produce the key-driver and outlook texts. For the detailed prompts employed in each stage, please refer to the Appendix \ref{appendix_prompt}.

\section{Experiment Results}

In this section, we demonstrate the experiment results on AIR framework. For all the experiments, we utilize a single NVIDIA RTX-3090 GPU as a computing device. We adopt an Adam \cite{adam} optimizer with learning rate of $10^{-4}$ for training.

\subsection{Dataset}
For the experiment, we introduce following two new multimodal time series datasets, in which the text modality is generated using our event-based data-refinement pipeline.

\textbf{Exchange Rate Dataset} To evaluate the performance of the AIR framework in the real-world forecasting environments with abundant text information, we experiment with time series forecasting problem in economic domain of exchange rate. 
As the time series data, we collect and utilize the daily exchange rates between the currencies of the top 20 GDP countries and the US dollar, as well as the US dollar index, covering the period from 2013 to 2024.
Additionally, we collect the stock indices of each country over the same period and use them as exogenous time series data\footnote{We collect time series data for exchange rate domain from Investing.com (https://www.investing.com).}
\footnote{Due to differences in public holidays, the exchange rate and stock index of Saudi Arabia are excluded.}. Among the collected time-series data, we use the US dollar index and the exchange rates between the US dollar and major currencies (European euro, Chinese yuan, Japanese yen, and British pound)\footnote{These major currencies constitute the Special Drawing Rights (SDR) valuation basket of International Monetary Fund (IMF) along with the US dollar.} as target variables. We display the more detailed information about time series data in Appendix \ref{appendix_exchange_rate}. For the text data, we utilize the Machine Readable News (MRN) data of London Stock Exchange Group (LSEG). First, we filter the data using news codes of MRN data, and then we apply our event-based text data refinement pipeline to further process the textual information and obtain key driver and outlook summary. 

\textbf{Crude Oil Price Dataset} In addition to exchange rate data, we utilize the crude oil price data to measure the performance of AIR in the real-world forecasting situation. 
For the crude oil price domain, we collect daily prices of Western Texas Intermediate (WTI) and European Brent crude oils spanning from 2013 to 2024 as target time series. We additionally collect exogenous time series data such as supply and demand of crude oil, and 
prices, supply, and demand of petroleum products, as well as futures prices of crude oil and petroleum products\footnote{We collect time series data for crude oil domain from EIA (https://www.eia.gov) and Investing.com (https://www.investing.com).}\footnote{Note that some exogenous time series data is provided in weekly basis, and we convert weekly data into daily data via linear interpolation.}. More detailed information about time series data can be found in Appendix \ref{appendix_crude_oil_price}.
For the text data, we again utilize LSEG MRN data. As done in exchange rate dataset, we filter the data using news code, and process the filtered data with our data refinement pipeline to obtain informative text data of key drivers and outlooks prepared for forecasting.

In the exchange rate and crude oil price data experiments, we set both the input length and prediction horizon to 20 days (4 weeks). The model performance is evaluated through a biweekly training experiment, where the model is trained using data collected up to every two-week point and then evaluated daily over the subsequent two-week period. The biweekly training experiment is conducted over a three-year period starting from the first week of 2022, and the average performance over the entire experimental period is measured.

\subsection{Models}

\begin{figure*}[h]
    \centering
    \includegraphics[width = 1.8\columnwidth]{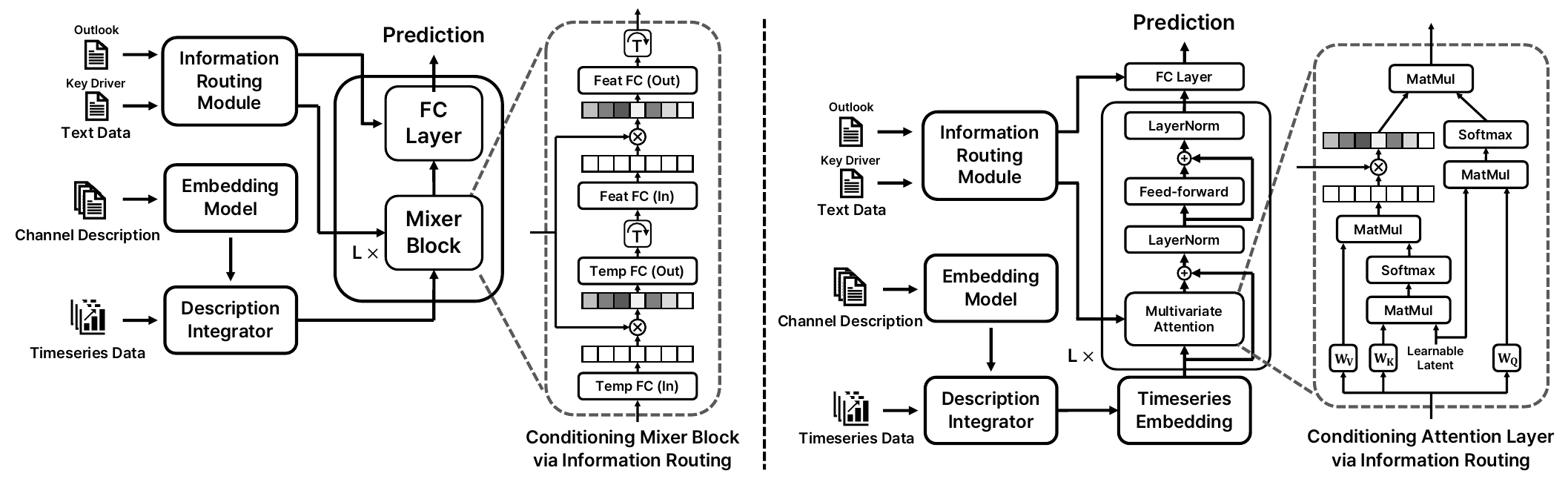}
    \caption{Adaptive Information Routing applied to TSMixer (left) and iTransformer (right). 
    }
    \label{fig4}
\end{figure*}
The proposed AIR framework is integrated with time-series forecasting models to endow them with multimodal forecasting capabilities. 
To demonstrate that AIR can be integrated into diverse forecasting models and enhance forecasting performance through the use of textual information, we apply AIR to various time-series forecasting models of varying network architectures and evaluate the resulting performance gains.
As base models for the AIR framework, we employ four architectures: TSMixer built upon fully connected layers, TCN that relies on on convolutional layers, and two attention-based models iTransformer and TimeXer.

In Figute \ref{fig4}, we illustrate the architecture of TSMixer (left) and iTransformer (right) integrated with the AIR framework. 
To incorporate domain-specific information into the routing process, we embed the descriptions of channels in time series data and integrate description embeddings with corresponding time series data, by concatenating the time series and description embedding and passing through an FC layer. 

TSMixer \cite{TSMixer} is a multivariate forecasting model based on MLPMixer \cite{MLPmixer} that relies exclusively on fully connected layers.
For TSMixer on the left side of Figure \ref{fig4}, information routing module generates the latent routing weights using key driver document for each FC layer in the mixer block. The information routing is also applied to the final predictor FC layer. 
The routing weights for the predictor layer are generated from the outlook text using a dedicated weight generator and vector quantization module, distinct from those used for the feature extractor.

On the right side of Figure \ref{fig4}, we display the architecture of iTransformer combined with the AIR framework. 
iTransformer is a transformer-based time series forecasting model that relies on the attention layer to model the inter-variable relationships in time series data.
In iTransformer, we apply AIR only to the attention layers. As done in TSMixer, we integrate description embedding with time series data, and apply time series embedding of iTransformer. For each Transformer block, we generate the routing weights from key driver test using information routing module and apply the weights to latent. We also generate the routing weights for the predictor using the outlook text and routing module, and apply the weights to the predictor.

Temporal Convolution Network (TCN) leverages convolutional networks to model temporal dependencies, consisting of a CNN based feature extractor and a fully connected predictor. 
For the TCN model, we apply AIR by assigning the routing weights generated from the key drivers to the convolution layers in the feature extractor, while the routing weights derived from the outlook text are applied to the predictor. 
TimeXer is a transformer-based time-series forecasting model that captures intra-variate and inter-variate dependencies through self-attention and cross-attention mechanisms, respectively.
For TimeXer model, we apply information routing of AIR to the cross attention part that models inter-variate dependencies by applying routing weight from key driver document. Consistent with the other base models, we generate the routing weights for the predictor using the outlook text and apply them accordingly.

\subsection{Results}
\begin{table*}[h]
        \centering
        \small
        \begin{tabular}{c|c|c|c|c|c|c|c|c}
            \toprule  
            \multicolumn{2}{c|}{Model}& WTI&Brent& USD Index&USD CNY&USD EUR&USD JPY&USD GBP\\
            \midrule
            \multirow{3}{*}{\raisebox{-0.5\height}{TSMixer}}&Vanilla&0.1006&0.1151&0.0766&0.0786&0.0730&0.1769&0.0666\\
            \cmidrule{2-9}
            &TimeMMD &0.1349&0.1611&0.1176&0.1088&0.1096&0.2266&0.0895\\
            \cmidrule{2-9}
            &AIR &\textbf{0.0667}&\textbf{0.0716}&\textbf{0.0530}&\textbf{0.0548}&\textbf{0.0518}&\textbf{0.1203}&\textbf{0.0465}\\
            \midrule
            \multirow{3}{*}{\raisebox{-0.5\height}{TCN}}&Vanilla&0.1311&0.1310&0.0909&0.1279&0.0885&0.2169&0.0754\\
            \cmidrule{2-9}
            &TimeMMD &0.1274&0.1453&0.1100&0.1305&0.1058&0.2237&0.0923\\
            \cmidrule{2-9}
            &AIR &\textbf{0.0833}&\textbf{0.0821}&\textbf{0.0553}&\textbf{0.0684}&\textbf{0.0544}&\textbf{0.1473}&\textbf{0.0483}\\
            \midrule
            \multirow{3}{*}{\raisebox{-0.5\height}{iTransformer}}&Vanilla&0.0893&0.1018&0.0745&0.0734&0.0766&0.1501&0.0722\\
            \cmidrule{2-9}
            &TimeMMD &0.1232&0.1299&0.0987&0.0773&0.0940&0.1577&0.0738\\
            \cmidrule{2-9}
            &AIR &\textbf{0.0757}&\textbf{0.0788}&\textbf{0.0590}&\textbf{0.0558}&\textbf{0.0598}&\textbf{0.1171}&\textbf{0.0473}\\
            \midrule
            \multirow{3}{*}{\raisebox{-0.5\height}{TimeXer}}&Vanilla&0.0880&0.0876&0.0645&0.0624&0.0680&0.1444&0.0543\\
            \cmidrule{2-9}
            &TimeMMD &0.1161&0.1197&0.0935&0.0729&0.0866&0.1672&0.0702\\
            \cmidrule{2-9}
            &AIR &\textbf{0.0726}&\textbf{0.0760}&\textbf{0.0533}&\textbf{0.0532}&\textbf{0.0526}&\textbf{0.1186}&\textbf{0.0473}\\
            \bottomrule
        \end{tabular}
	\caption{Crude oil price and exchange rate forecasting experiment results.}
        \label{table1}
    \end{table*}

In Table \ref{table1}, we display the experimental results with exchange rate and crude oil price data. We compare the multimodal performance of AIR with  AIR 
We compare the performance gains achieved by multimodal forecasting using AIR with those of the existing multimodal forecasting baseline of TimeMMD \cite{TimeMMD}.
Both approaches can be integrated with existing time-series forecasting models to enable multimodal forecasting. For the TimeMMD baseline in our experiments, we embed the two documents of key driver and outlook using the same embedding model employed in AIR, concatenate the resulting embeddings, project them through an MLP network, and then combine the projected prediction with the prediction of unimodal base model. 

As we can see in Table \ref{table1}, AIR consistently yields performance improvements across all targets and all basemodels, whereas TimeMMD baseline shows little or no performance gain in most cases. For models based on feed-forward architectures such as TSMixer and TCN, AIR achieves substantial performance improvements. In TSMixer, AIR lowers the MSE loss by an average of 31.97\%, and in TCN, AIR achieves an average reduction of 37.96\% across all targets.
Even in Transformer-based models with stronger forecasting capabilities, AIR achieves the notable performance improvements. AIR reduces the loss of iTransformer by 23\% and that of TimeXer by 16.61\% across 7 targets in crude oil price and exchange rate domain.

In Table \ref{table2}, we compare the performance of the proposed AIR framework with the prompt-based multimodal forecasting method, Cik \cite{cik}. For CiK experiment, we utilized three large language models—GPT-4o and GPT-4o-mini, which are closed-source, and EXAONE-3.5-32B, an open-source alternative. 
We utilize the prompt provided in the CiK paper, modifying its context section to align with our text data and guide the model to generate forecasts conditioned on the key-driver and outlook information.
Due to the knowledge cutoff date of LLMs, we evaluate the performance only for the year 2024. The table \ref{table2} demonstrates that, while prompt-based forecastings of CiK show performance levels similar to unimodal forecasting models, integrating AIR with a forecasting backbone yields superior results, surpassing both the unimodal models and the CiK method.

\begin{table*}[h]
        \centering
        \small
        \begin{tabular}{c|c|c|c|c|c|c|c|c}
            \toprule  
            \multicolumn{2}{c|}{Model}& WTI&Brent& USD Index&USD CNY&USD EUR&USD JPY&USD GBP\\
            \midrule
            \multirow{3}{*}{\raisebox{-1.5ex}{CiK}} &GPT-4o&0.0373&0.0364&0.0333&0.0168&0.0413&0.0784&0.0299\\
            \cmidrule{2-9}
            &GPT-4o-mini &0.0421&0.0420&0.0363&0.0159&0.0346&0.0820&0.0298\\
            \cmidrule{2-9}
            &EXAONE-3.5-32B &0.0481&0.0475&0.0839&0.0236&0.0390&0.0837&0.0320\\
            \midrule
            \multirow{4}{*}{\raisebox{-2.0ex}{Unimodal}} &TSMixer &0.0463&0.0425&0.0308&0.0147&0.0333&0.0870&0.0260\\
            \cmidrule{2-9}
            &TCN &0.0441&0.0465&0.0459&0.0272&0.0419&0.1086&0.0292\\
            \cmidrule{2-9}
            &iTransformer &0.0388&0.0380&0.0345&0.0181&0.0361&0.0807&0.0259\\
            \cmidrule{2-9}
            &TimeXer &0.0308&0.0325&0.0295&0.0157&0.0305&0.0732&0.0217\\
            \midrule
            \multirow{4}{*}{\raisebox{-2.0ex}{AIR}} &TSMixer &0.0286&\textbf{0.0301}&0.0265&0.0138&0.0262&\textbf{0.0614}&0.0191\\
            \cmidrule{2-9}
            &TCN &0.0353&0.0359&0.0293&\textbf{0.0130}&0.0297&0.0616&0.0242\\
            \cmidrule{2-9}
            &iTransformer &\textbf{0.0278}&0.0302&0.0271&0.0138&0.0259&0.0661&\textbf{0.0179}\\
            \cmidrule{2-9}
            &TimeXer &0.0279&0.0295&\textbf{0.0259}&0.0145&\textbf{0.0256}&0.0633&0.0187\\
            \bottomrule
        \end{tabular}
	\caption{Comparison of Adaptive Information Routing and prompt-based baseline.}

        \label{table2}
    \end{table*}
\subsection{Ablation Study}
Table \ref{table3} presents the ablation study results of the AIR framework utilizing the TSMixer architecture.
In the table, “+AIR on F” indicates that information routing based on key drivers is applied to the mixer layers of the feature extractor, while “+AIR on P” denotes that information routing derived from the outlook text is further applied to the predictor. The results demonstrate a consistent enhancement in forecasting performance as more textual information is utilized and additional information routing components are integrated into the model. The last row labeled "+VQ" denotes the performance of full AIR framework including the application of vector quantization. The gains achieved by incorporating vector quantization suggest that structuring the routing weights via quantization stabilizes the information routing process and thereby enhancing forecasting accuracy.
\begin{table*}[h]
        \centering
        \small
        \begin{tabular}{c|c|c|c|c|c|c|c|c}
            \toprule  
            \multicolumn{2}{c|}{Model}& WTI&Brent& USD Index&USD CNY&USD EUR&USD JPY&USD GBP\\
            \midrule
            \multirow{4}{*}{\raisebox{-2.0ex}{TSMixer}}&Vanilla&0.1006&0.1151&0.0766&0.0786&0.0730&0.1769&0.0666\\
            \cmidrule{2-9}
            &+AIR on F &0.0861&0.0905&0.0843&0.0695&0.0732&0.1596&0.0616\\
            \cmidrule{2-9}
            &+AIR on P&0.0786&0.0779&0.0667&0.0665&0.0664&0.1436&0.0540\\
            \cmidrule{2-9} 
            &+VQ &\textbf{0.0667}&\textbf{0.0716}&\textbf{0.0530}&\textbf{0.0548}&\textbf{0.0518}&\textbf{0.1203}&\textbf{0.0465}\\
            \bottomrule
        \end{tabular}
	\caption{Ablation experiment results on TSMixer integrated with AIR.}

        \label{table3}
    \end{table*}

\section{Discussion and Future Works}
This paper focuses on the fundamental differences between time series data and text data, and proposes a multimodal time series forecasting framework of AIR reflecting the differences. 
While the experiment results demonstrate the validity of our approach, there remains substantial room for further improvement in effectively leveraging multimodal data. 
Recent work suggests that textual data contains various types of information such as historical, future covariate, and causal information \cite{cik}. As our AIR framework which accounts for the fundamental differences between textual and time series information successfully achieves performance improvements in multimodal time series forecasting, leveraging textual data with diverse characteristics in an appropriate manner is expected to lead to even better forecasting performance. For example, using text data containing historical or future information as input and text data containing covariate or causal information as a controller can lead to greater performance improvement compared to using only one of the two types of approach.


\section{Conclusion}
In this paper, we introduce the Adaptive Information Routing (AIR) framework, a novel approach for multimodal time series forecasting. AIR framework modifies the behavior of multivariate forecasting model adaptively to the text data by adjusting connection between input and output nodes in fully-connected layers. By decomposing the layer in model into two distinct layers with an intervening latent,
AIR can adaptively control the information path through which the time series information flows by weighting the latent between the decomposed layers.  
We also introduce an event-based text data refinement pipeline that leverages the large language models to organize and refine the scattered information in the news data into a more informative form. Using on our data refinement pipeline, we introduce two new multimodal time series dataset in crude oil price and exchange rate domain. Experiment results on our benchmark show that the proposed AIR framework is able to utilize the text information to improve the forecasting performance in real world time series forecasting problem.
\bibliography{reference}
\bibliographystyle{icml2025}

\newpage
\appendix
\onecolumn
\appendix
\section{Dataset Details}
In this section, we provide the details about the datasets we utilize for evaluation in the main paper.

\subsection{Exchange Rate Data}\label{appendix_exchange_rate}
In Table \ref{tableA1}, we display the detailed information on time series data in our exchange rate dataset. As a base time series, we utilize the exchange rates between the US Dollar and the currencies of top 20 GDP countries.
Under the assumption that each country's stock market index reflects its economic conditions, and can thus serve as an indicator for estimating changes in the value of its currency, we utilize stock indices as exogenous time series data.
We collect time series data of exchange rates and stock market indices from https://www.investing.com/. For the exchange rates, we utilize exchange rates from USD to AUD, BRL, CAD, CHF, CNY, EUR, GBP, HKD, IDR, INR, JPY, KRW, MXN, RUB, TRY. For stock market indices, we utilize indices of Netherlands, Turkey, Brazil, France, Germany, European Union, United Kingdom, Italy, Hong Kong, Spain, Indonesia, South Korea, Russia, India, Japan, United States of America, Australia, Mexico, Canada, Switzerland, and China. Due to differing national holidays of different countries, there are timestamps for which values are missing in some time series. In such cases, the missing values are filled using linear interpolation. As a result, we obtain a time series dataset with a length of 3,131 and 37 variables.

For the text data, we first filter the document in LSEG MRN dataset using the news code included in metadata of MRN document data. Using the data refinement pipeline outlined in Section \ref{section:data_pipeline}, we process the textual data and daily key-driver and outlook texts based on the news content contained in a 20-day rolling window. We embed the documents into fixed-size vectors with length 4096 using the LGAI-embedding-preview embedding model.

\begin{table*}[h]
        \centering
        \scriptsize
        \begin{tabular}{l|l|c|c}
            \toprule  
            time series&Description&Frequency &Source\\
            \midrule
            USD Index (Target)&US Dollar Index&Daily & INV\\
            USD AUD &Exchange Rate of US Dollar to Australian Dollar&Daily & INV\\
            USD BRL &Exchange Rate of US Dollar to Brazilian Real&Daily & INV\\
            USD CAD &Exchange Rate of US Dollar to Canada Dollar&Daily & INV\\
            USD CHF &Exchange Rate of US Dollar to Swiss Franc&Daily & INV\\
            USD CNY (Target) &Exchange Rate of US Dollar to Chinese Yuan&Daily & INV\\
            USD EUR (Target) &Exchange Rate of US Dollar to European Euro&Daily & INV\\
            USD GBP (Target) &Exchange Rate of US Dollar to British Pound&Daily & INV\\
            USD HKD &Exchange Rate of US Dollar to Hong Kong Dollar&Daily & INV\\
            USD IDR &Exchange Rate of US Dollar to Indonesian Rupiah&Daily & INV\\
            USD INR &Exchange Rate of US Dollar to Indian Rupee&Daily & INV\\
            USD JPY (Target) &Exchange Rate of US Dollar to Japanese Yen&Daily & INV\\
            USD KRW &Exchange Rate of US Dollar to Korean Won&Daily & INV\\
            USD MXN &Exchange Rate of US Dollar to Mexican Peso&Daily & INV\\
            USD RUB &Exchange Rate of US Dollar to Russian Ruble&Daily & INV\\
            USD TRY &Exchange Rate of US Dollar to Turkish Lira&Daily & INV\\
            AEX &Netherlands Stock Market Index&Daily & INV\\
            BIST 100 &Turkish Stock Market Index&Daily & INV\\
            Bovespa &Brazilian Stock Market Index&Daily & INV\\
            CAC 40 &France Stock Market Index &Daily & INV\\
            DAX &Germany Stock Market Index&Daily & INV\\
            Euro Stoxx 50 &European Union Stock Market Index&Daily & INV\\
            FTSE 100 &British Stock Market Index&Daily & INV\\
            FTSE MIB &Italian Stock Market Index&Daily & INV\\
            Hang Seng &Hong Kong Stock Market Index&Daily & INV\\
            IBEX 35 &Spainish Stock Market Index&Daily & INV\\
            IDX Composite &Indonesian Stock Market Index&Daily & INV\\
            KOSPI &Korean Stock Market Index&Daily & INV\\
            MOEX Russia Index &Russian Stock Market Index&Daily & INV\\
            Nifty 50 &Indian Stock Market Index&Daily & INV\\
            Nikkei 225 &Japanese Stock Market Index&Daily & INV\\
            S\&P 500 &United States of America Stock Market Index&Daily & INV\\
            S\&P ASX 200 &Australian Stock Market&Daily & INV\\
            S\&P/BMV IPC &Mexican Stock Market Index&Daily & INV\\
            S\&P/TSX Composite &Canada Stock Market Index&Daily & INV\\
            SMI &Swiss Stock Market Index&Daily & INV\\
            Shanghai Composite &Chinese Stock Market Index&Daily & INV\\

            \bottomrule
        \end{tabular}
	\caption{Detailed information on time series data in exchange rate dataset. INV denotes Investing.com (https://www.investing.com).}

        \label{tableA1}
    \end{table*}

\subsection{Crude Oil Price Data}\label{appendix_crude_oil_price}
In Table \ref{tableA2}, we provide the details on time series data in crude oil price dataset. For crude oil price forecasting, we utilize information on the prices and futures prices of crude oil and petroleum products, crude oil production, demand for petroleum products, refinery input and output volumes, and inventories and import/export volumes of crude oil and petroleum products. For factors other than crude oil prices and futures prices, we choose to use the corresponding indicators from the United States—one of the largest crude oil producers and the largest consumer market—as representative values for these variables. The U.S. Energy Information Administration (EIA) systematically provides information on crude oil and energy markets, including domestic crude oil production, imports, refining volumes, exports, and inventories and we utilize this data to construct our dataset.
Among the time series, price-related information is available on a daily basis, while the other variables are available on a weekly basis. For the weekly time series, we convert them into daily time series through a linear interpolation. As a result, we obtain a time series dataset with a length of 3,129\footnote{Due to differences in the process of filling missing values based on daily time series, the crude oil price dataset lacks values for December 25, 2013, and December 25, 2024, resulting in a time series length that differs from the exchange rate dataset.} and 76 variables.

As done in exchange rate dataset, we filter the document data using the news code, and process the document data applying the data refinement pipeline described in Section \ref{section:data_pipeline}. The daily key-driver and outlook texts obtained from the data refinement pipeline are then embedded into the fixed-size vectors with length 4096 with the LGAI-embedding-preview embedding model. 
\begin{table*}[h!]
        \centering
        \scriptsize
        \begin{tabular}{l|c|c|c}
            \toprule  
            time series&Frequency &Category &Source\\
            \midrule
            Cushing, OK WTI Spot Price FOB (Target)&Daily &Crude Oil Price& EIA\\
            Europe Brent Spot Price FOB (Target)&Daily&Crude Oil Price & EIA\\
            Heating Oil Futures Price&Daily &Futures Price& INV\\
            Brent Oil Futures Price&Daily &Futures Price& INV\\
            Gasoline RBOB Futures Price&Daily &Futures Price& INV\\
            Crude Oil WTI Futures Price&Daily&Futures Price & INV\\
            New York Harbor Conventional Gasoline Regular Spot Price FOB &Daily & Petroleum Products Price& EIA\\
            U.S. Gulf Coast Conventional Gasoline Regular Spot Price FOB &Daily & Petroleum Products Price& EIA\\
            Los Angeles Reformulated RBOB Regular Gasoline Spot Price&Daily & Petroleum Products Price& EIA\\
            New York Harbor No. 2 Heating Oil Spot Price FOB&Daily & Petroleum Products Price& EIA\\
            New York Harbor Ultra-Low Sulfur No 2 Diesel Spot Price&Daily & Petroleum Products Price& EIA\\
            U.S. Gulf Coast Ultra-Low Sulfur No 2 Diesel Spot Price&Daily & Petroleum Products Price& EIA\\
            Los Angeles, CA Ultra-Low Sulfur CARB Diesel Spot Price&Daily & Petroleum Products Price& EIA\\
            U.S. Gulf Coast Kerosene-Type Jet Fuel Spot Price FOB&Daily & Petroleum Products Price& EIA\\
            Mont Belvieu, TX Propane Spot Price FOB&Daily & Petroleum Products Price& EIA\\
            U.S. Field Production of Crude Oil &Weekly &Crude Oil Production& EIA\\
            U.S. Product Supplied of Petroleum Products&Weekly &Petroleum Products Demand & EIA\\
            U.S. Product Supplied of Finished Motor Gasoline&Weekly &Petroleum Products Demand& EIA\\
            U.S. Product Supplied of Kerosene-Type Jet Fuel&Weekly&Petroleum Products Demand & EIA\\
            U.S. Product Supplied of Distillate Fuel Oil &Weekly &Petroleum Products Demand& EIA\\
            U.S. Product Supplied of Residual Fuel Oil&Weekly &Petroleum Products Demand& EIA\\
            U.S. Product Supplied of Propane and Propylene&Weekly&Petroleum Products Demand & EIA\\
            U.S. Product Supplied of Other Oils&Weekly &Petroleum Products Demand& EIA\\
            U.S. Refiner Net Input of Crude Oil&Weekly  &Refining Input& EIA\\
            U.S. Refiner and Blender Net Input of Gasoline Blending Components&Weekly&Refining Input & EIA\\
            U.S. Refiner and Blender Net Input of Fuel Ethanol&Weekly &Refining Input& EIA\\
            U.S. Refiner and Blender Net Production of Kerosene-Type Jet Fuel &Weekly &Refining Production& EIA\\
            U.S. Refiner and Blender Net Production of Distillate Fuel Oil &Weekly &Refining Production& EIA\\
            U.S. Refiner and Blender Net Production of Residual Fuel Oil&Weekly &Refining Production& EIA\\
            U.S. Refiner, Blender, and Gas Plant Net Production of Propane and Propylene&Weekly &Refining Production& EIA\\
            U.S. Oxygenate Plant Production of Fuel Ethanol&Weekly &Refining Production& EIA\\
            U.S. Ending Stocks of Crude Oil and Petroleum Products &Weekly&Stock & EIA\\
            U.S. Ending Stocks excluding SPR of Crude Oil and Petroleum Products&Weekly&Stock & EIA\\
            U.S. Ending Stocks of Crude Oil&Weekly &Stock& EIA\\
            U.S. Ending Stocks excluding SPR of Crude Oil&Weekly&Stock & EIA\\
            U.S. Ending Stocks excluding SPR and including Lease Stock of Crude Oil&Weekly &Stock& EIA\\
            U.S. Crude Oil Stocks in Transit (on Ships) from Alaska&Weekly &Stock& EIA\\
            U.S. Ending Stocks of Crude Oil in SPR&Weekly &Stock& EIA\\
            U.S. Ending Stocks of Total Gasoline&Weekly &Stock& EIA\\
            U.S. Ending Stocks of Fuel Ethanol&Weekly &Stock& EIA\\
            U.S. Ending Stocks of Kerosene-Type Jet Fuel&Weekly &Stock& EIA\\
            U.S. Ending Stocks of Distillate Fuel Oil&Weekly &Stock& EIA\\
            U.S. Ending Stocks of Residual Fuel Oil&Weekly &Stock& EIA\\
            U.S. Ending Stocks of Propane and Propylene&Weekly&Stock & EIA\\
            U.S. Ending Stocks of Other Oils&Weekly&Stock & EIA\\
            U.S. Ending Stocks of Unfinished Oils&Weekly  &Stock& EIA\\
            U.S. Ending Stocks of Kerosene&Weekly &Stock& EIA\\
            U.S. Ending Stocks of Asphalt and Road Oil&Weekly &Stock& EIA\\
            U.S. Ending Stocks of NGPLs/LRGs (Excluding Propane/Propylene) &Weekly&Stock & EIA\\
            U.S. Imports of Crude Oil and Petroleum Products&Weekly &Import& EIA\\
            U.S. Imports of Crude Oil&Weekly &Import& EIA\\
            U.S. Commercial Crude Oil Imports Excluding SPR&Weekly &Import& EIA\\
            U.S. Crude Oil Imports by SPR&Weekly &Import& EIA\\
            U.S. Crude Oil Imports for SPR by Others&Weekly &Import& EIA\\
            U.S. Imports of Total Petroleum Products&Weekly &Import& EIA\\
            U.S. Imports of Total Gasoline&Weekly &Import& EIA\\
            U.S. Imports of Fuel Ethanol&Weekly &Import& EIA\\
            U.S. Imports of Kerosene-Type Jet Fuel&Weekly &Import& EIA\\
            U.S. Imports of Distillate Fuel Oil&Weekly &Import& EIA\\
            U.S. Imports of Residual Fuel Oil&Weekly &Import& EIA\\
            U.S. Imports of Propane and Propylene&Weekly&Import & EIA\\
            \midrule
            \multicolumn{4}{r}{\textit{(continued on next page)}} \\
            \bottomrule
        \end{tabular}

    \end{table*}

\begin{table*}[h!]
        \centering
        \scriptsize
        \begin{tabular}{l|c|c|c}
            \toprule  
            time series&Frequency &Category &Source\\    
            \midrule
            U.S. Imports of Other Oils (Excluding Fuel Ethanol)&Weekly &Import& EIA\\
            U.S. Imports of Kerosene&Weekly &Import& EIA\\
            U.S. Imports of Liquefied Petroleum Gasses Less Propane/Propylene &Weekly &Import& EIA\\
            U.S. Net Imports of Crude Oil and Petroleum Products&Weekly &Import& EIA\\
            U.S. Net Imports of Crude Oil &Weekly &Import& EIA\\
            U.S. Net Imports of Total Petroleum Products&Weekly&Import & EIA\\
            U.S. Exports of Crude Oil and Petroleum Products&Weekly &Export& EIA\\
            U.S. Exports of Crude Oil&Weekly &Export& EIA\\
            U.S. Exports of Total Petroleum Products&Weekly &Export& EIA\\
            U.S. Exports of Total Motor Gasoline&Weekly &Export& EIA\\
            U.S. Exports of Kerosene-Type Jet Fuel &Weekly &Export& EIA\\
            U.S. Exports of Total Distillate&Weekly &Export& EIA\\
            U.S. Exports of Residual Fuel Oil&Weekly &Export & EIA\\
            U.S. Exports of Propane&Weekly &Export& EIA\\
            U.S. Exports of Other Oils &Weekly&Export & EIA\\

            \bottomrule
        \end{tabular}
	\caption{Detailed information on time series data in crude oil price dataset. INV denotes Investing.com (https://www.investing.com/) and EIA denotes U.S. Energy Information Administration (https://www.eia.gov/) }

        \label{tableA2}
    \end{table*}

\subsection{Prompt for Large Language Model}\label{appendix_prompt}
In Section \ref{section:data_pipeline}, we describe the event-based text data refinement pipeline that processes textual data and extracts the forecast-relevant information using large language models. Our data refinement pipeline consists of three main stages: the news-wise filter and summary, daily event extraction, and window-wise insight generation. In this section, we present detailed explanations of the individual stages of our data-refinement pipeline, along with the prompts used at each stage. 

\begin{figure*}[h]
    \centering
    \includegraphics[width = 0.9\columnwidth]{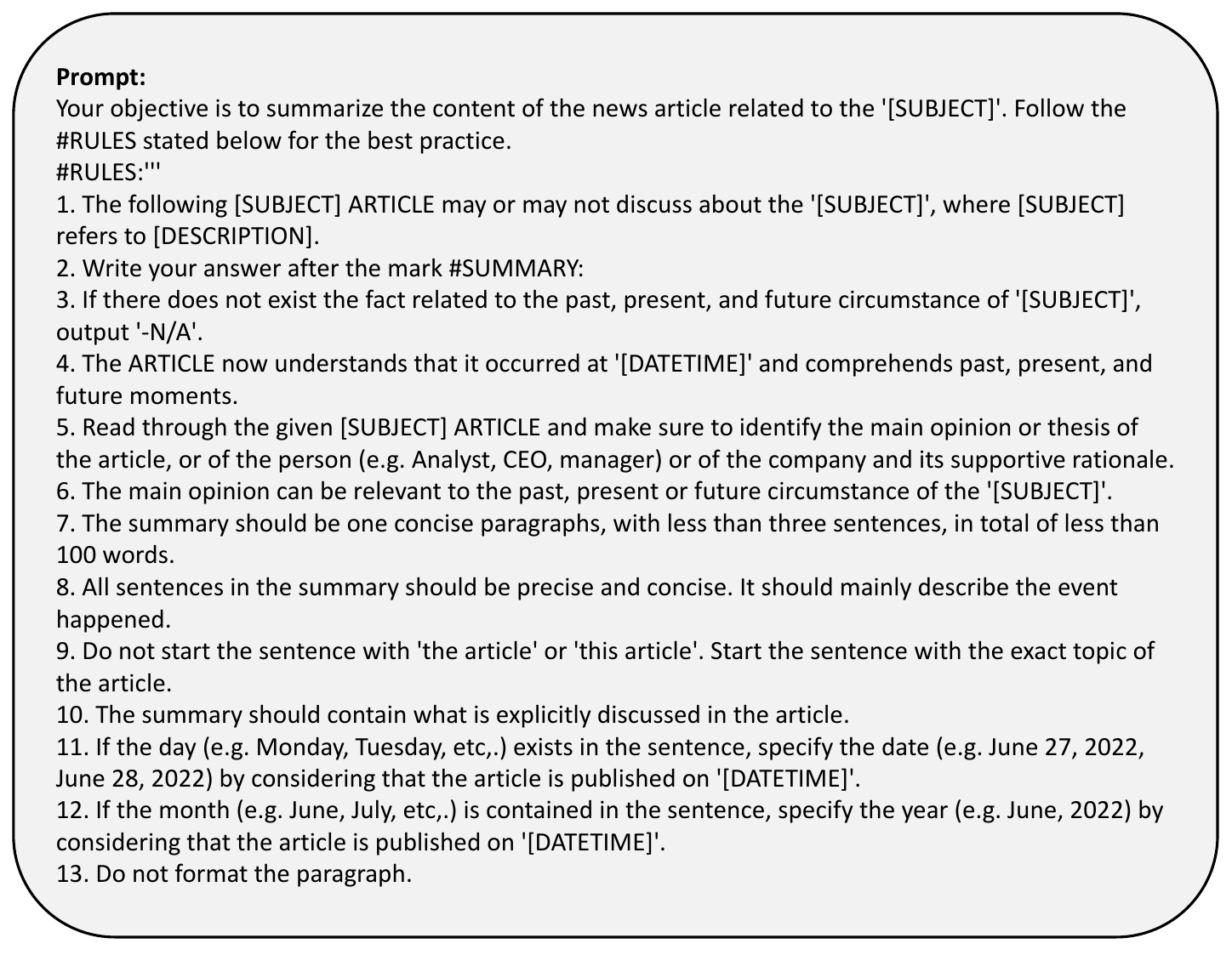}
    \caption{Prompt for summarizing and filtering documents. 
    }
    \label{figA5_1}
\end{figure*}

In Figure \ref{figA5_1}, we display the prompt used in filter and summary stage. Using this prompt, we obtain the news-summary set by collecting all outputs except those labeled as N/A. For the exchange rate dataset, we set SUBJECT as 'exchange rate' and DESCRIPTION as 'foreign exchange market'. For the crude oil price data, we use SUBJECT of 'crude oil' and DESCRIPTION of 'crude oil market'. We obtain DATETIME information from the datetime information provided with document in LSEG MRN dataset. 

\begin{figure*}[h]
    \centering
    \includegraphics[width = 0.9\columnwidth]{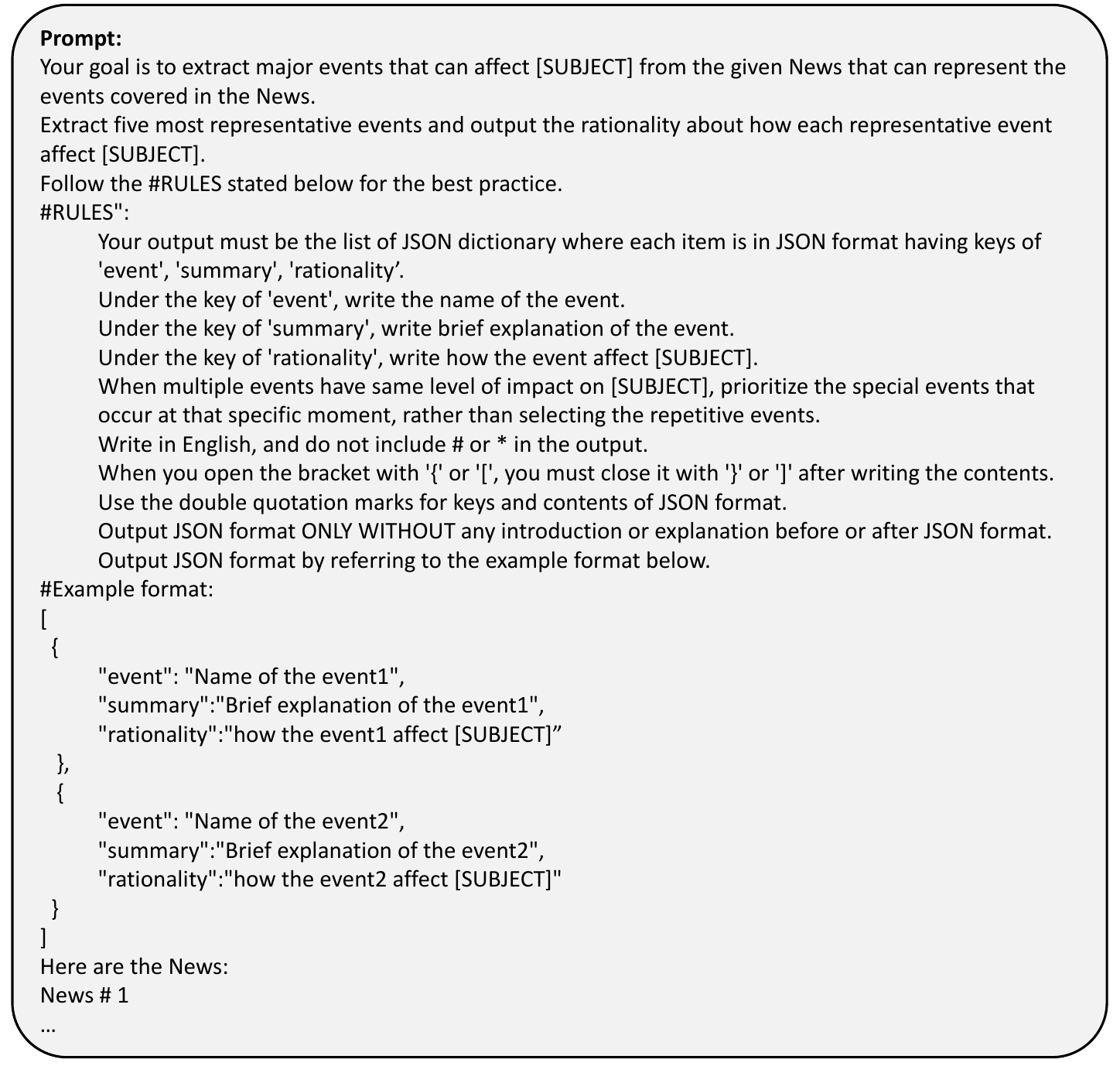}
    \caption{Prompt for daily event extraction. 
    }
    \label{figA5_2}
\end{figure*}

\begin{figure*}[h]
    \centering
    \includegraphics[width = 0.9\columnwidth]{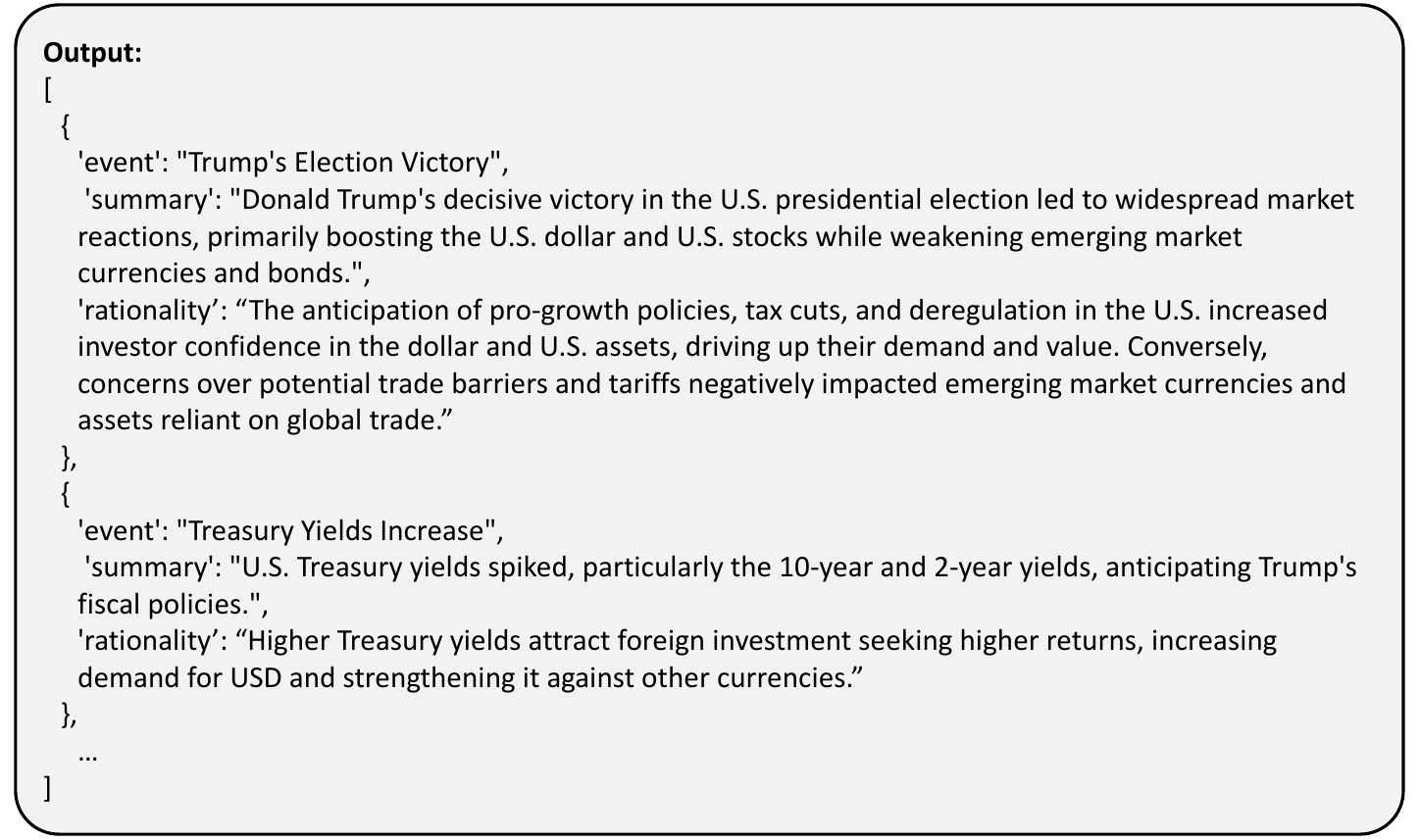}
    \caption{Output example for daily event extraction. 
    }
    \label{figA5_2_2}
\end{figure*}

In Figure \ref{figA5_2}, we demonstrate the prompt for the daily event extraction stage. Using this prompt, we extract the five most representative events for each day from the daily news-summary set organized in the form of a JSON-formatted list.
For the exchange rate and crude oil price dataset, we use SUBJECT of 'exchange rate' and 'crude oil', respectively. We illustrate an example of the daily events extracted using the prompt in Figure \ref{figA5_2} for November 5, 2024, the date on which the outcome of the U.S. presidential election was reported in Figure \ref{figA5_2_2}. The example demonstrates that the LLM correctly extracts the outcome of the U.S. presidential election as a key event, accompanied by an appropriate summary and rationality.

\begin{figure*}[h]
    \centering
    \includegraphics[width = 0.9\columnwidth]{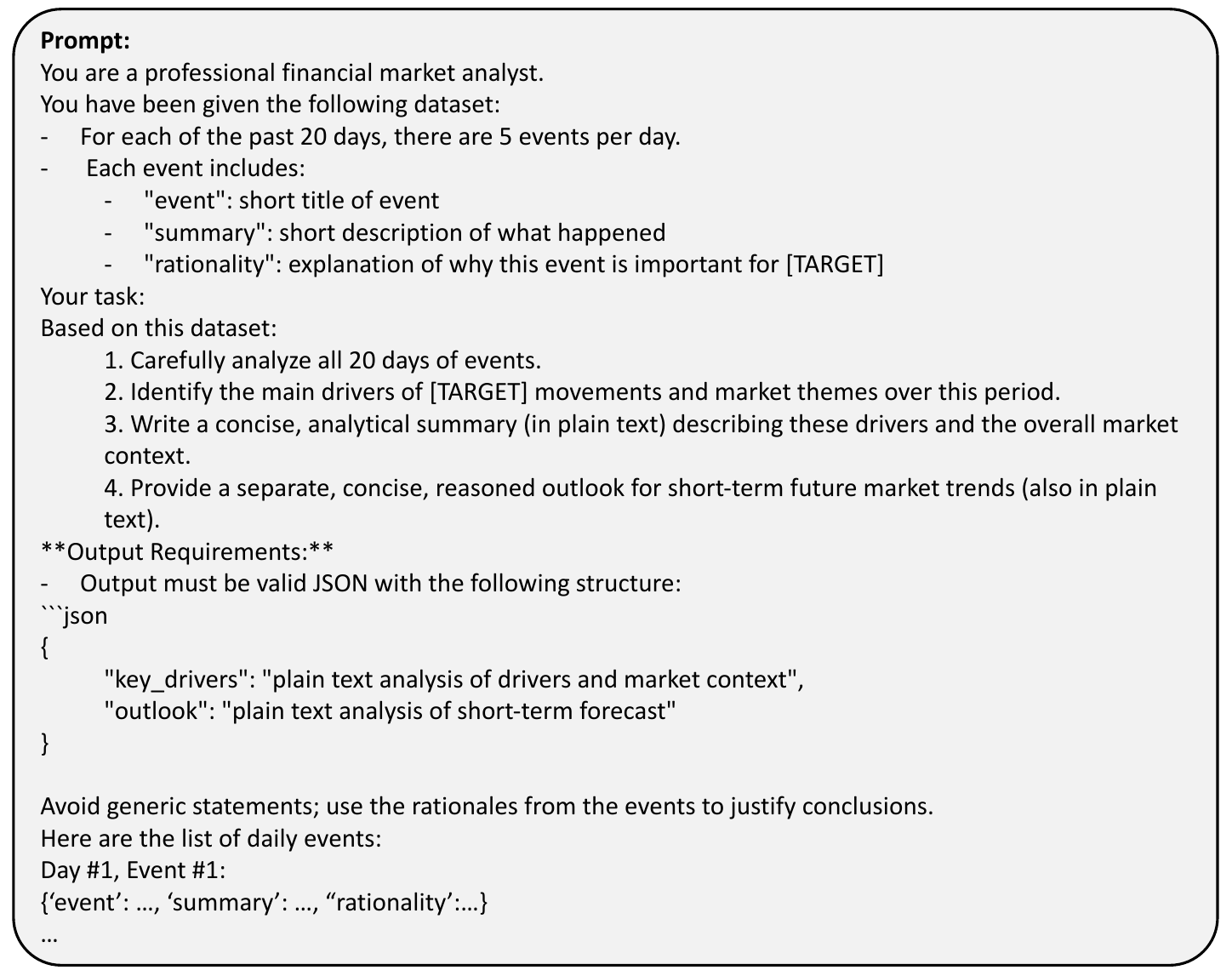}
    \caption{Prompt for window-wise insight generation. 
    }
    \label{figA5_3}
\end{figure*}

\begin{figure*}[h]
    \centering
    \includegraphics[width = 0.9\columnwidth]{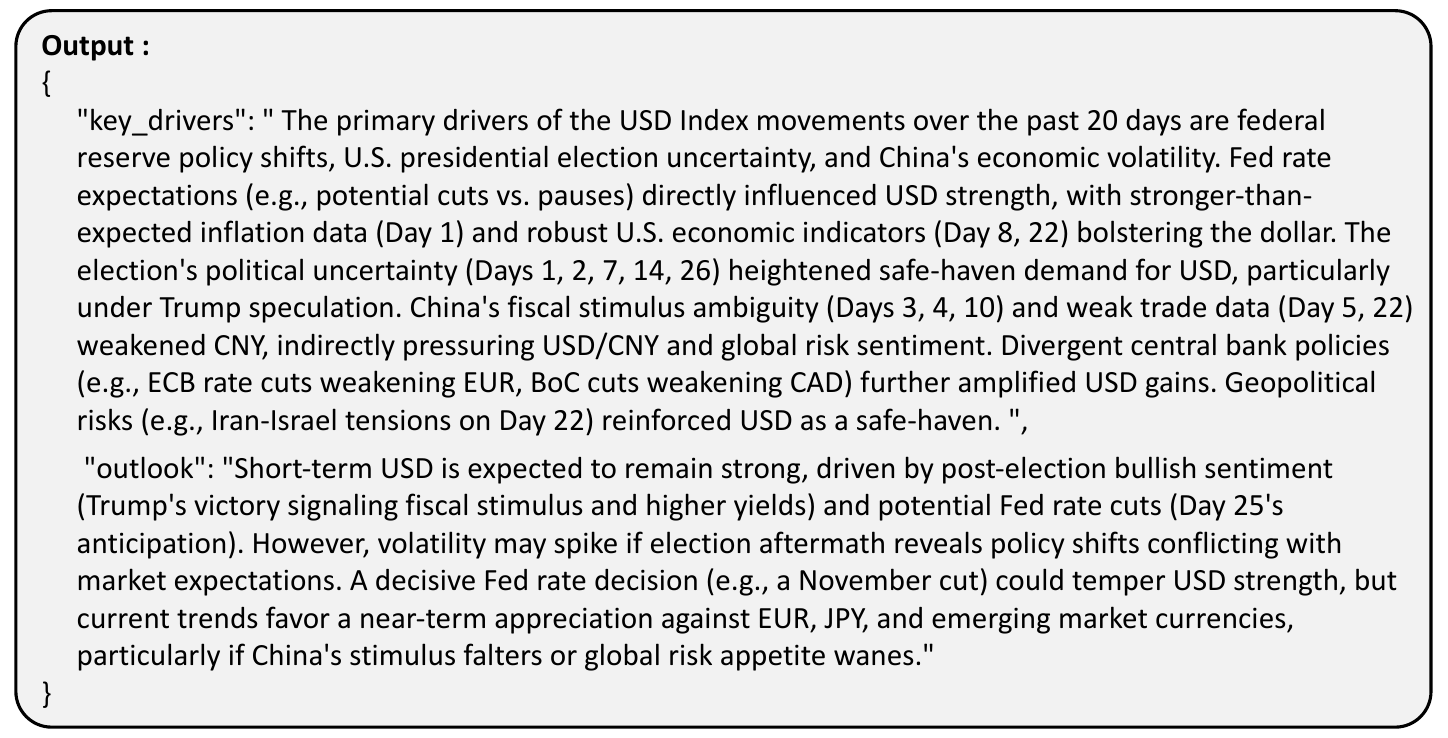}
    \caption{Output example for window-wise insight generation. 
    }
    \label{figA5_3_2}
\end{figure*}

In Figure \ref{figA5_3}, we present the prompt used in window-wise insight generation stage. With this prompt, we provide the LLM with the events spanning the 20-day window and generate insight texts distilled from the event information that are relevant for forecasting. The generated insight text is composed of two parts: the key driver, capturing current market dynamics along with the analytical factors that explain them, and the outlook, which conveys short-term forecasts of forthcoming market movements. For the TARGET in the prompt, we use the text explanation of each target, as displayed in Table \ref{tableA3}. Figure \ref{figA5_3_2} displays the window-wise insight generation output for the U.S. Dollar Index corresponding to November 5, 2024. In Figure \ref{figA5_3_2}, the key-driver analysis correctly identifies relevant elements—including election-related uncertainty, Federal Reserve policy considerations, and China’s economic environment. Furthermore, the outlook articulates a persuasive scenario regarding potential shifts in the U.S. Dollar Index following the election outcome.

\begin{table*}[h!]
        \centering
        \scriptsize
        \begin{tabular}{c|c|l}
            \toprule  
            Dataset&Target &Description\\    
            \midrule
            \multirow{2}{*}{\raisebox{-0.5ex}{Crude Oil Price}} &WTI&West Texas Intermediate Crude Oil Price\\
            \cmidrule{2-3}
            &Brent&European Brent Crude Oil Price\\
            \midrule
            \multirow{5}{*}{\raisebox{-3.5\height}{Exchange Rate}} &USD Index&United States of America Dollar Index\\
            \cmidrule{2-3}
            &USD CNY&Exchange Rate of United States of America Dollar to Chinese Yuan\\
            \cmidrule{2-3}
            &USD EUR&Exchange Rate of United States of America Dollar to European Euro\\
            \cmidrule{2-3}
            &USD JPY&Exchange Rate of United States of America Dollar to British Pound\\
            \cmidrule{2-3}
            &USD GBP&Exchange Rate of United States of America Dollar to Japanese Yen\\
            \bottomrule
        \end{tabular}
	\caption{Target descriptions for window-wise insight generation}
        \label{tableA3}
    \end{table*}



\section{Further Details on Biweekly Training Experiment}
In exchange rate, crude oil price dataset, we evaluate the model performance through biweekly training experiment. In this section, we provide more details about weekly/biweekly training experiments. For the ordering of weeks within a given year, we use the week index defined according to the ISO 8601 standard. Under this standard, there exists 52 weeks in 2022 to 2024. For each experiment in the biweekly training setting, the starting point of the test data is specified using a week index in the format 'year-week'. 
In the biweekly training setting, the evaluation period for the test data spans 29 days, during which ten 20-day forecasts are performed—each starting from a weekday between the Monday of the given week and the Friday of the following week. Note that due to missing dates caused by holidays, the starting points of the evaluation period in the test data may not align exactly with Monday through Friday, or with Monday to the following Friday, as intended.
For the training and validation, we utilize the all available data before the Monday of the given week. The model is trained for up to 40 epochs using all available data. Given the complexity of the task and to avoid underfitting, we configure the training and validation sets to be identical and continue training until the validation loss no longer improved. We utilize the mean squared error (MSE) as the training loss. For each experiment, we conduct 3 experiments with 3 different random seeds and use the MSE averaged across 3 random seeds as the performance metric.

\section{Visualization for Multimodal Forecasting with AIR}
\begin{figure*}[h]
    \centering
    \includegraphics[width = 1.0\columnwidth]{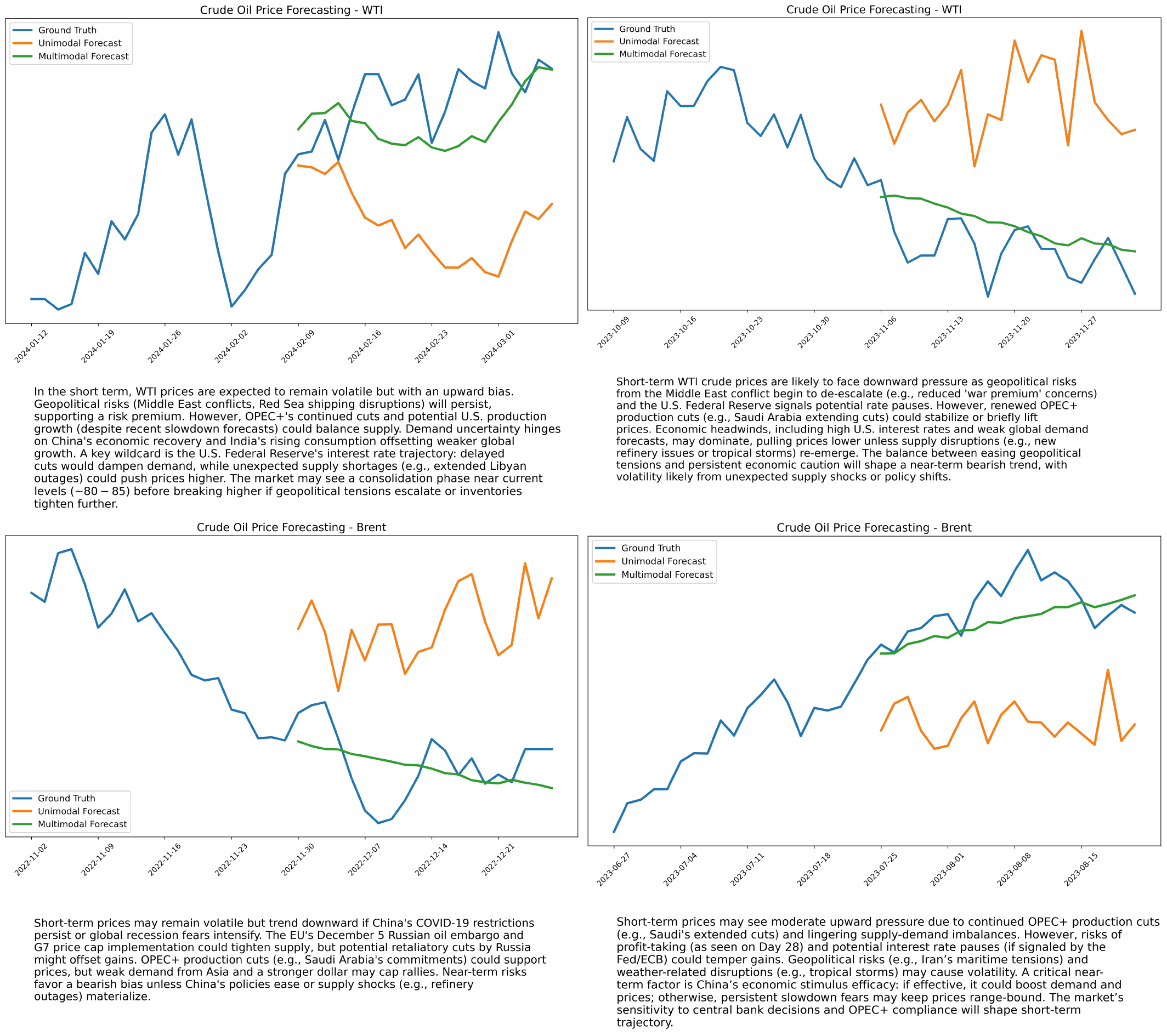}
    \caption{Visualization of crude oil price forecasting results from forecasting model with and without AIR. 
    }
    \label{figA6_1}
\end{figure*}

In Figure \ref{figA6_1}, we demonstrate the forecasting results of forecasting models combined with AIR, in crude oil price dataset with the outlook text generated from our event-based text data refinement pipeline. We can see that the multimodal forecasting model combined with AIR outperforms the unimodal forecasting model. We can also observe that the information provided through the outlook text effectively guides the model,steering its forecasts toward a more appropriate and accurate outcome. We visualize the forecasting results of unimodal and multimodal forecasting from the exchange rate dataset experiment, along with the outlook text. These results again demonstrate that the multimodal forecasting model appropriately utilizes textual information, enabling it to surpass the performance of the unimodal forecasting model.


\begin{figure*}[h]
    \centering
    \includegraphics[width = 1.0\columnwidth]{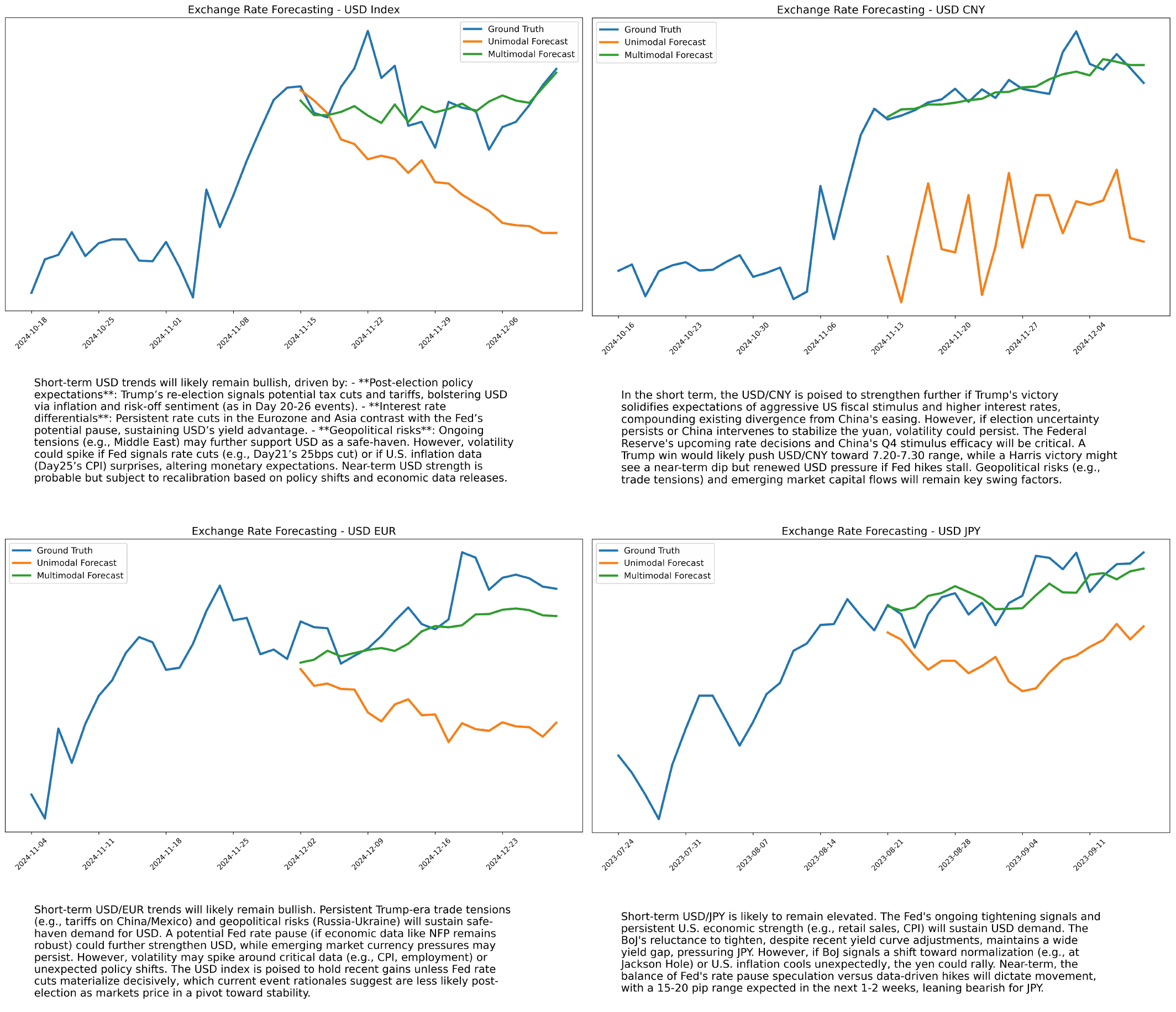}
    \caption{Visualization of exchange rate forecasting results from forecasting model with and without AIR. 
    }
    \label{figA6_2}
\end{figure*}

\end{document}